\documentclass{article} % For LaTeX2e
\usepackage[T1]{fontenc}
\usepackage[preprint]{colm2026_conference}

\usepackage{microtype}
\usepackage{hyperref}
\usepackage{url}
\usepackage{booktabs}
\usepackage{amsmath}
\usepackage{algorithm}
\usepackage{algorithmic}
\usepackage{graphicx}
\usepackage{amsthm}
\usepackage{xcolor}
\usepackage{subcaption}
\usepackage[table]{xcolor}
\definecolor{ourblue}{RGB}{221,235,247}
\usepackage{wrapfig}
\usepackage{tabularx}
\usepackage{enumitem}
\usepackage[most]{tcolorbox}
\tcbuselibrary{breakable,listings,skins}
\usepackage{xspace}
\usepackage{textcomp}

\newcommand{\method}{\textsc{RCSD}\xspace}

\makeatletter
\renewcommand{\@fnsymbol}[1]{%
  \ensuremath{%
    \ifcase#1\or
      \dagger\or
      \ddagger\or
      \mathsection\or
      \mathparagraph\or
      \|\or
      **\or
      \dagger\dagger\or
      \ddagger\ddagger
    \else
      \@ctrerr
    \fi
  }%
}
\makeatother

\usepackage{lineno}

\definecolor{darkblue}{rgb}{0, 0, 0.5}
\hypersetup{colorlinks=true, citecolor=darkblue, linkcolor=darkblue, urlcolor=darkblue}

\title{Rethinking Reward Supervision: Rubric-Conditioned Self-Distillation
}

\author{Siyi Gu \\
Yale University \\
\texttt{siyi.gu@yale.edu} \\
\And
Jialin Chen \\
Yale University \\
\texttt{jialin.chen@yale.edu} \\
\And
Sophia Zhou \\
Yale University \\
\texttt{sophia.zhou@yale.edu} \\
\And
Arman Cohan$^\dagger$ \\
Yale University \\
\texttt{arman.cohan@yale.edu} \\
\And
Rex Ying\thanks{Equal advising} \\
Yale University \\
\texttt{rex.ying@yale.edu}
}

\begin{document}

\ifcolmsubmission
\linenumbers
\fi

\maketitle

\begin{abstract}
Post-training of reasoning language models is commonly driven by on-policy distillation and reinforcement learning with verifiable rewards. Distillation often relies on chain-of-thought annotations that are expensive to obtain and may themselves be noisy, incomplete, or partially incorrect; even when the final solution is correct, an imperfect rationale can interfere with learning. Reinforcement learning with verified rewards, on the other hand, typically compresses evaluative feedback into a scalar signal, obscuring which aspects of a response should be improved. We propose \textbf{Rubric-Conditioned Self-Distillation}, a framework that incorporates rubrics as structured, fine-grained feedback for on-policy self-distillation. Our method conditions the teacher model on criterion-level rubrics and uses it to provide token-level guidance on the student’s own sampled trajectories. This design avoids treating a single reference rationale as the sole supervision target. Instead, rubrics specify what a strong response should satisfy, enabling more fine-grained credit assignment over the reasoning process than scalar reward optimization. We instantiate this framework with a two-stage pipeline that first learns to generate task-specific rubrics and then trains a rubric-guided reasoner. We evaluate on a diverse suite of science reasoning benchmarks and results show that rubric-conditioned self-distillation effectively converts rubric-level criteria into token-level guidance over the reasoning process, surpassing GRPO by 1.0 points and OPSD by 0.9 points on average. Code available:  \url{https://github.com/carriegu0818/RCSD}. 
\end{abstract}

\section{Introduction}
Recent advances in large language models (LLMs) have led to substantial progress in reasoning, problem-solving, and instruction following. Reinforcement learning has been particularly effective in domains such as mathematics and code generation, where final outcomes can be automatically verified. However, Group Relative Policy Optimization (GRPO) objective typically optimizes sparse outcome-level rewards: the model is rewarded only after generating a complete response, for example, based on whether the final answer is correct or whether execution succeeds \citep{shao2024deepseekmath}. While effective in verifiable settings, such supervision provides little information about \emph{why} a trajectory succeeds or fails, creating a persistent credit-assignment bottleneck in online learning \citep{hubotter2026sdpo, zhao2026opsd}.

One natural way to enrich this supervision is through \emph{rubrics}. Rather than scoring a response holistically, rubrics decompose quality into explicit criteria, yielding a more structured and interpretable representation of what makes an answer strong \citep{gunjal2025rubrics,zhang2025chasing}. Recent work has shown that rubric-based evaluation can extend post-training beyond strictly verifiable tasks by supplying richer judgments than binary correctness alone \citep{gunjal2025rubrics}. Yet in most existing works, rubric information enters training only through the \emph{reward}: criterion-level judgments are aggregated into a single scalar score and then optimized with RL-style updates on the entire trajectory. As a result, the rich information inside such textual rubric feedback is largely discarded during optimization.

A recent line of work addresses sparse outcome rewards by replacing them with dense teacher supervision. In on-policy distillation (OPD) and on-policy self-distillation (OPSD), the student learns from its own sampled trajectories while a teacher provides token-level guidance along those rollouts \citep{agarwal2024policy,xu2024speculative,zhao2026opsd,hubotter2026sdpo,ye2026opcd}. These methods alleviate the mismatch of off-policy imitation and provide a denser learning signal than final-answer rewards. However, existing approaches typically construct the teacher from stronger-model outputs. Such supervision is therefore tied to particular privileged trajectories, which may not cleanly expose the underlying dimensions along which a response should be evaluated. These trajectories represent only one valid reasoning trace rather than the underlying dimensions that define a strong response. In this sense, rationale-based supervision can over-specify \emph{how} an answer should be produced, without cleanly identifying \emph{what properties} the answer should satisfy.

\begin{wrapfigure}{r}{0.5\columnwidth}
    \vspace{-0.8em}
    \centering
    \includegraphics[width=0.48\columnwidth]{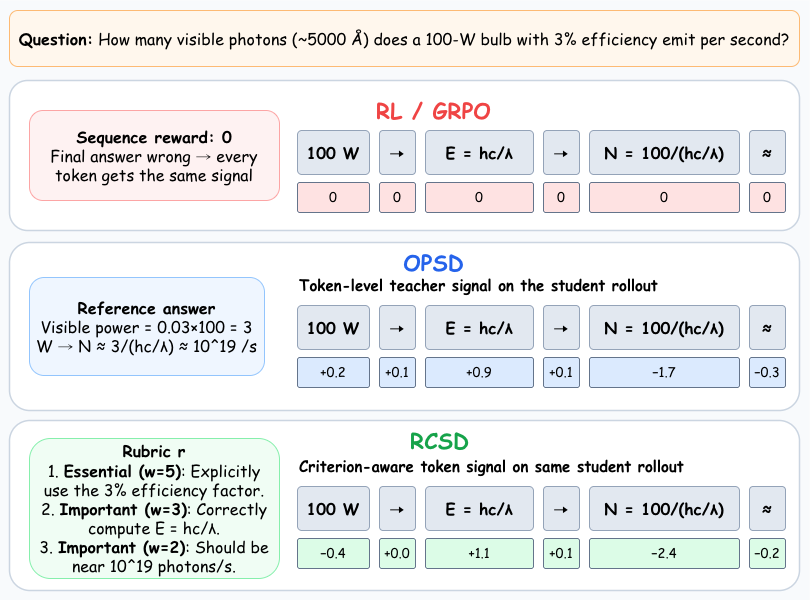}
    \caption{Illustration of how optimization signals differ between RL/OPSD/\method with incorrect student trajectory. }
    \label{fig:rgsd_token_example}
    \vspace{-1.0em}
\end{wrapfigure}

In this work, we introduce \emph{\textbf{R}ubric-\textbf{C}onditioned \textbf{S}elf-\textbf{D}istillation} (RCSD), a post-training framework that uses rubrics as privileged teacher-side supervision for on-policy self-distillation. Our key idea is that rubrics should not only score responses after generation; they should shape token-level learning during optimization.  Instead of collapsing rubric feedback into a scalar reward, we condition the teacher on criterion-level rubric information and distill its token-level guidance on the student’s own sampled trajectories. The resulting training signal is simultaneously \emph{criterion-aware}, \emph{on-policy}, and \emph{token-level}: it preserves distinctions across evaluation dimensions, operates on student-generated rollouts rather than fixed off-policy traces, and provides dense guidance without reducing feedback to a single number.  Figure~\ref{fig:rgsd_token_example} illustrates the optimization signal difference on the same incorrect trajectory: RL assigns one reward to the full sequence, OPSD provides dense supervision toward a reference trajectory, and \method provides dense rubric-conditioned feedback that preserves correct steps while penalizing the specific local error.

We operationalize our idea with a two-stage pipeline. We first train a rubric generator to amortize instance-specific evaluation criteria from privileged supervision, and then train a reasoner with rubric-conditioned teacher guidance. More broadly, we reframe rubrics as a structured supervision interface for model self-improvement, especially in hard-to-verify and open-ended tasks where high-quality responses are not fully captured by automatic verification or scalar outcome rewards. Across diverse reasoning benchmarks, \method achieves the best overall average (70.6), surpassing GRPO by 1.4 points and OPSD by 0.9 points. Notably, the gains are pronounced on scientific and rubric-based reasoning tasks, where response quality is poorly captured by scalar outcome-level rewards alone.

\section{Method}
\begin{figure}[t]
    \centering
    \includegraphics[width=\linewidth]{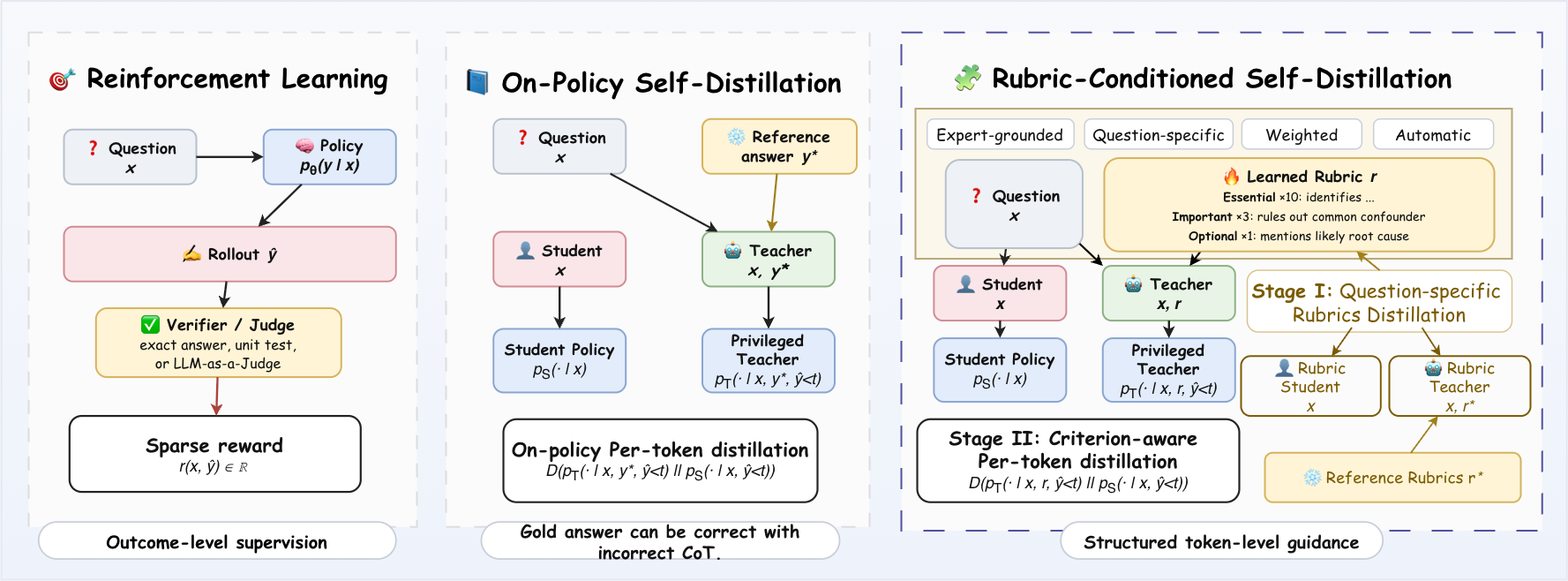}
    \caption{\method uses rubrics as privileged teacher-side supervision for on-policy self-distillation. In contrast to RL which compresses feedback into a scalar reward, and OPSD which conditions the teacher on a reference answer, \method learns question-specific rubrics in Stage I and reuses them in Stage II to induce structured token-level guidance on the student’s own reasoning trajectory.}
    \label{fig:main}
\end{figure}

%\subsection{Method Overview}
We propose to preserve the fine-grained, structured feedback during optimization by using learned rubrics as privileged teacher-side supervision for on-policy self-distillation. Rather than compressing rubric feedback into a single number, we expose it to a privileged teacher, which then provides dense token-level guidance on the student's own sampled trajectories. Figure~\ref{fig:main} situates our method relative to two standard alternatives. Reinforcement learning applies outcome-level supervision through a sparse scalar reward. On-policy self-distillation replaces this with token-level teacher guidance, but typically conditions the teacher on a privileged reference answer. In contrast, our method redefines this supervision interface: instead of conditioning the privileged teacher on a single reference trajectory, we condition it on a rubric that specifies criterion-level properties of a strong response.
%We instantiate this idea in two stages. Stage~I learns a rubric generator that predicts instance-specific evaluation criteria from the question alone. Stage~II trains a reasoner using rubric-conditioned teacher guidance on on-policy student rollouts. 

\subsection{Preliminaries}

We use $p_T$ and $p_S$ to denote the teacher and student distributions, respectively.

\textbf{Off-Policy Distillation}~\citep{hinton2015distilling} trains a student to imitate trajectories generated by a teacher. In its most general form, the objective can be written as
\begin{equation}
\mathcal{L}_{\mathrm{off}}
  =
  \mathbb{E}_{x \sim \mathcal{D},\, y \sim p_T(\cdot \mid x)}
  \left[
  \sum_{t=1}^{|y|}
  D\!\left(
  p_T(\cdot \mid x, y_{<t})
  \,\|\,
  p_S(\cdot \mid x, y_{<t})
  \right)
  \right],
\end{equation}

where $D(\cdot\|\cdot)$ denotes a divergence between teacher and student sequence distributions. 
%In practice, large language model distillation is commonly instantiated as supervised fine-tuning on teacher-generated responses,
% \begin{equation}
% \mathcal{L}_{\mathrm{SFT}}
% =
% -\mathbb{E}_{x \sim \mathcal{D},\; y \sim p_T(\cdot \mid x)}
% \left[
% \sum_{t=1}^{|y|}
% \log p_S(y_t \mid x, y_{<t})
% \right].
% \end{equation}
Off-policy distillation provides dense token-level supervision, but it suffers from a distribution mismatch: the student is trained on teacher-generated prefixes, whereas at inference time it must condition on its own generated prefixes, leading to compounding errors and degraded performance.

\textbf{On-Policy Distillation (OPD)}~\citep{agarwal2024policy,gu2024minillm} addresses this mismatch by sampling trajectories from the student rather than the teacher. Given an input $x$, the student first generates an on-policy rollout $\hat{y} \sim p_S(\cdot \mid x).$
The teacher and student are then compared along the student's own trajectory, yielding the objective
\begin{equation}
\mathcal{L}_{\mathrm{OPD}}
=
\mathbb{E}_{x \sim \mathcal{D},\; \hat{y} \sim p_S(\cdot \mid x)}
\left[
\frac{1}{|\hat{y}|}
\sum_{t=1}^{|\hat{y}|}
D\!\left(
p_T(\cdot \mid x, \hat{y}_{<t})
\,\|\, 
p_S(\cdot \mid x, \hat{y}_{<t})
\right)
\right].
\end{equation}
However, the on-policy distillation method still heavily relies on token-level imitation of a teacher distribution, which often encourages the student to follow a single preferred response, ignoring the space of valid reasoning paths.

\textbf{On-Policy Self-Distillation (OPSD)}~\citep{zhao2026opsd,hubotter2026sdpo}
 refers to the setting in which the teacher and student are derived from the same underlying model rather than from two separately trained models. In the on-policy self-distillation setting, a single model instantiates both a student policy and a privileged teacher policy. Given a reasoning dataset $S=\{(x,z)\}$, where $z$ denotes privileged information such as a gold solution, a reference answer, or other side information, the student observes only the base input $x$ and generates an on-policy response $\hat{y} \sim p_S(\cdot \mid x)$, while the teacher is conditioned on privileged information $z$ unavailable to the student at inference time. The OPSD objective is
\begin{equation}
\mathcal{L}_{\mathrm{OPSD}}
  =
  \mathbb{E}_{(\textcolor{red}{x,z})\sim \textcolor{red}{\mathcal{S}},\; \hat{y}\sim p_S(\cdot\mid x)}
  \left[
  \frac{1}{|\hat{y}|}
  \sum_{t=1}^{|\hat{y}|}
  D\!\left(
  p_T(\cdot \mid x, \textcolor{red}{z}, \hat{y}_{<t})
  \,\|\,
  p_S(\cdot \mid x, \hat{y}_{<t})
  \right)
  \right].
\end{equation}
While OPSD further introduces privileged information to guide learning, it suffers from conditioning the teacher on a specific reference solution contained in $z$. This can be restrictive for reasoning tasks, where solution quality is better characterized by satisfying a set of criteria rather than a sole target. These limitations motivate a more flexible supervision interface that provides structured, multi-dimensional criterion-level guidance.

% \ac{Consider using colors so reader can more easily see the diff with OPD. Example:}

% \[
%   \mathcal{L}_{\mathrm{OPSD}}
%   =
%   \mathbb{E}_{(\textcolor{red}{x,z})\sim \textcolor{red}{\mathcal{S}},\; \hat{y}\sim p_S(\cdot\mid x)}
%   \left[
%   \frac{1}{|\hat{y}|}
%   \sum_{t=1}^{|\hat{y}|}
%   D\!\left(
%   p_T(\cdot \mid x, \textcolor{red}{z}, \hat{y}_{<t})
%   \,\|\,
%   p_S(\cdot \mid x, \hat{y}_{<t})
%   \right)
%   \right]
% \]  

% \ac{end-of-comment}

\subsection{Motivation: Beyond Reward Optimization and Reference-Conditioned Distillation}

We position \method against two common paradigms for improving reasoning models: reward-based optimization and reference-conditioned distillation.

\paragraph{(1) Reward optimization requires sparse external judgment.}
GRPO has been highly effective in verifiable domains, where correctness can be directly checked by exact-match answers or unit tests~\citep{shao2024deepseekmath,guo2025deepseek,chollet2025arc,jain2024livecodebench}. Recent work extends this paradigm to hard-to-verify or non-verifiable domains by using LLM-as-a-Judge rewards~\citep{li2026rubrichub,gunjal2025rubrics}. However, this extension still reduces supervision to sparse scalar reward signals, which provide limited information about which intermediate reasoning steps should be improved. It also introduces an additional external evaluator, increasing inference and training cost, and may further amplify reward bias from the judge model through its preferences, calibration errors, or inconsistent interpretation of rubrics. In contrast, \method does not require a separate reward model or judge during distillation. Instead, we provide rubrics directly to the teacher model and let the teacher generate rubric-conditioned reasoning on its own outputs. This turns rubrics into a structured supervision interface, allowing the student to learn from dense token-level teacher guidance rather than optimizing against sparse scalar reward signals.

\paragraph{(2) Reference-conditioned distillation is path-specific.}
OPSD conditions the teacher on a single reference trajectory, which can induce path-specific supervision. When the student deviates from this trajectory, even slightly, the teacher signal may encourage global revision rather than localized correction. Empirically, we observe that OPSD trajectories often recompute the same intermediate quantities or revise earlier steps without new information, leading to long and redundant reasoning chains. This suggests that OPSD provides token-level supervision, but lacks explicit criterion-level credit assignment. Our proposed method, \method, addresses this limitation by conditioning the teacher on rubric criteria instead of a single reference path, yielding supervision that is both on-policy and criterion-aware.

\subsection{Problem Setup}
Let \(x\) denote an input question. We consider two structured outputs associated with \(x\): a rubric \(r\) and a answer \(y\). A rubric is a structured set of question-specific evaluation criteria, $r = \{c_1, \dots, c_K\}$,
where each criterion \(c_k\) contains a title, a natural language description, and an importance weight $\in \textit{Essential}, \textit{Important}, \textit{Optional}, or \textit{Pitfall}$. Conceptually, a rubric serves as an intermediate supervision interface that specifies what constitutes a good solution and provide richer and more interpretable information on how to guide the model optimization. A high-quality rubric encourages the model to focus on satisfying high-level criteria rather than imitating a specific reasoning path.

% A reasoning trajectory is a token sequence
% $y = (y_1,\dots,y_T)$,
% representing the model's generated solution.

Our goal is to learn a model that produces high-quality reasoning trajectories under instance-specific evaluation criteria. Such criteria are useful during training, but are unavailable to the student at inference time. We therefore treat rubrics as \emph{privileged supervision}: they are provided to the teacher during training and distilled into the student through on-policy token-level guidance. Since distilling high-quality rubrics could be costly, we automate the rubric generation process and further factorize learning into two sequential stages: 
\begin{enumerate}[leftmargin=0.5cm, nosep]
    \item \textbf{Stage I: Learning a rubric generator.} We train a model to predict a rubric \(r\) conditioned on the question \(x\), amortizing instance-specific evaluation criteria into a reusable form.
    \item \textbf{Stage II: Rubric-conditioned reasoning.} We train a reasoner to generate trajectories that satisfy rubric criteria, using the rubric as structured guidance during optimization.
\end{enumerate}

% Training proceeds on-policy in both stages. The student samples trajectories from its current policy, while a privileged teacher, instantiated from the same model but conditioned on additional information,  provides dense token-level supervision.

\subsection{Stage I: Learning a Rubric Generator}

A practical challenge in rubric-based training is that high-quality instance-specific rubrics are expensive to obtain. Our first stage therefore amortizes privileged supervision into a standalone rubric generator. During training, the teacher is allowed to view both the question and a reference answer, while the student must learn to infer an appropriate rubric from the question alone.

\paragraph{Student policy.}
The student rubric generator observes only the question:
$ p_S^R(r \mid x).$

\paragraph{Teacher policy.}
The teacher rubric generator receives privileged access to the question and reference answer \(y^\star\):
$p_T^R(r \mid x, y^\star).$
The reference answer is not used as a supervision target to be copied directly; rather, it serves as a privileged context that helps the teacher infer how to generate a correct response trajectory within its own distribution.

\paragraph{On-policy rubric distillation.}
Given a sampled rubric rollout $\hat r \sim p_S^R(\cdot \mid x)$, we train the student to match the teacher's next-token distribution along the student's own rubric trajectory:
\begin{equation}
\label{eq:rubric_loss}
\mathcal{L}_{\text{rubric}}
=
\mathbb{E}_{\hat r \sim p_S^R(\cdot \mid x)}
\left[
\sum_{t=1}^{|\hat r|}
D_{\mathrm{KL}}
\Big(
p_T^R(\cdot \mid x, y^\star, \hat r_{<t})
\;\|\;
p_S^R(\cdot \mid x, \hat r_{<t})
\Big)
\right].
\end{equation}

This objective preserves the key advantage of on-policy self-distillation: the student is trained on its own sampled prefixes rather than teacher-generated ones. At the same time, the teacher can inject privileged information from the reference answer to shape the rubric-generation process. As a result, Stage~I distills evaluation structure into a rubric generator that can produce question-specific criteria without requiring privileged inputs at test time.

\subsection{Stage II: Rubric-Conditioned Reasoning}

Given a rubric, the second stage trains a reasoner to generate answers that better satisfy instance-specific criteria. The key design choice is that the rubric is not merely appended as auxiliary prompt text for the student. Instead, it is provided as a privileged context to the teacher, which uses it to deliver criterion-aware token-level guidance on the student's own rollout.

\paragraph{Student policy.}
The student reasoner observes only the question:
$p_S^Y(y \mid x).$

\paragraph{Teacher policy.}
The teacher reasoner receives the question together with rubric feedback:
$p_T^Y(y \mid x, r).$
Here \(r\) denotes the learned rubric from the Stage~I rubric generator. Conditioning the teacher on \(r\) allows the training signal to reflect multiple dimensions of response quality, rather than a single scalar reward or a single reference trajectory.

\paragraph{On-policy rubric-conditioned distillation.}
Given an on-policy student rollout
$\hat y \sim p_S^Y(\cdot \mid x),$
we minimize
\begin{equation}
\label{eq:reason_loss}
\mathcal{L}_{\text{reason}}
=
\mathbb{E}_{\hat y \sim p_S^Y(\cdot \mid x)}
\left[
\sum_{t=1}^{|\hat y|}
D_{\mathrm{KL}}
\Big(
p_T^Y(\cdot \mid x, r, \hat y_{<t})
\;\|\;
p_S^Y(\cdot \mid x, \hat y_{<t})
\Big)
\right]
.
\end{equation}

This objective highlights the core advantage of on-policy distillation—dense supervision on student-generated trajectories. Rather than learning from a reference rationale or a scalarized reward, the student is guided by a teacher conditioned on criterion-level rubric information. Its key importance is that criterion-level structure is retained in the teacher signal itself, so optimization can distinguish different dimensions of partial correctness rather than compressing them into a single undifferentiated score.

% \paragraph{Why rubrics provide a better supervision signal.}
Reference-conditioned distillation supervises the student with one specific
solution trajectory, which can introduce path-specific bias when multiple
derivations are valid. Rubrics instead specify the criteria that a correct
solution must satisfy, inducing a criterion-aware teacher distribution over
many valid reasoning paths rather than a single reference path. This distinction naturally aligns with forward KL distillation. Minimizing
$D_{\mathrm{KL}}(p_T(\cdot \mid x,r)\,\|\,p_S(\cdot \mid x))$ encourages the
student to cover the support of the rubric-conditioned teacher distribution,
preserving probability mass on alternative solutions that satisfy the same
criteria. In contrast, more mode-seeking objectives may concentrate on dominant
teacher modes and underrepresent valid but less likely derivations. Thus,
forward KL provides a principled objective for transferring dense,
criterion-level guidance while maintaining diversity across valid reasoning
paths.

\subsection{Training Procedure}
\label{sec:training_procedure}
Algorithm~\ref{alg:method} summarizes the full training procedure. Stage~I amortizes expensive instance-specific criteria into a rubric generator, and Stage~II uses those criteria to shape teacher-side token-level correction on student rollouts. The resulting framework preserves criterion-level structure, remains on-policy, and better utilize this rich structured information in textual rubrics rather than scalar rewards.

\begin{algorithm}[H]
\caption{\method: Rubric-Conditioned On-Policy Self-Distillation}
\label{alg:method}
\small
\begin{algorithmic}[1]
\REQUIRE Training set \(\mathcal{D}=\{(x,  r^\star, y^\star)\}\); rubric generator \(p_S^R\); reasoner \(p_S^Y\) 

\STATE \textbf{Stage I: Learn rubric generator}
\FOR{each training example \(x\) with reference answer \(y^\star\)}
    \STATE Sample rubric trajectory \(\hat r \sim p_S^R(\cdot \mid x)\)
    \STATE Compute teacher and student token distributions along \(\hat r\)
    \STATE Update rubric generator by minimizing \(\mathcal{L}_{\text{rubric}}\)
\ENDFOR

\STATE \textbf{Stage II: Train rubric-conditioned reasoner}
\FOR{each training example \(x\)}
    \STATE Obtain rubric \(\hat r\) from \(\hat r \sim p_S^R(\cdot \mid x)\)
    \STATE Sample answer rollout \(\hat y \sim p_S^Y(\cdot \mid x)\)
    \STATE Compute teacher and student token distributions along \(\hat y\)
    \STATE Update reasoner by minimizing \(\mathcal{L}_{\text{reason}}\)
\ENDFOR
\end{algorithmic}
\end{algorithm}

% \paragraph{Complexity and overhead.}
% Compared with OPSD, \method adds a one-time Stage~I cost for training the rubric generator and a small teacher-side overhead in Stage~II from conditioning on rubrics. The Stage~II update itself remains the same on-policy token-level distillation procedure on student rollouts, so the main practical overhead is rubric generation plus slightly longer teacher context.
\method introduces a new supervision interface for post-training, by using rubrics as privileged structured guidance for on-policy self-distillation. Overall, Stage~I learns \emph{what} a strong response should satisfy, while Stage~II learns \emph{how} to realize those criteria along the student's own trajectory. Prior rubric-based RL typically uses rubrics only for outcome-level scoring, while standard OPSD preserves token-level optimization but ties supervision to a single privileged answer. \method differs from both by letting fine-grained rubric feedback directly guide on-policy learning.

% \rex{complexity and potential overhead analysis}

% \rex{after 2.3, 2.4 and 2.5, it would be useful to emphasize again the advantage, and provide motivation for the method design in terms of criteria improvement (2.3), token-level reward (2.4), co-evolution (2.5) or whatever you think is relevant. This is where we can emphasize novelty a bit more for reviewers to appreciate our contribution.}
\section{Experiments}
% \sg{EXPERIMENT: 1) main table: science reasoning 2) generalization to medicine 3) ablation on generated/GT rubric 4) model size. (5) model response length v.s. training steps - probably not able to finish) Consider drop math reasoning since aime and math500 results don't show improvement.
% }

\label{sec:experiments}

Our experiments are designed to answer the following questions: (1) whether rubric-guided token-level feedback improves both verifiable and open-ended reasoning performance compared with scalar-reward and reference-conditioned distillation baselines; (2) whether the resulting gains persist across model scales and remain robust on out-of-domain benchmarks; and (3) whether learned rubrics can approach the effectiveness of reference rubrics, and how sensitive \method is to rubric quality.

 %, and (4) how \method compares with prior post-training methods in both reasoning performance and sample efficiency.

\textbf{Data Construction.} We follow the two-stage design described in Section~\ref{sec:training_procedure} to construct rubric generation data and reasoning data. For rubric learning, we construct our training set based on \textsc{RaR-Science}~\citep{gunjal2025rubrics} and \textsc{RubricHub}~\citep{li2026rubrichub} for reference rubric supervision. For reasoner training, we additionally take \texttt{natural\_reasoning~\citep{yuan2025naturalreasoning}} and filter out entries with empty reference answers. The final dataset contains
approximately 10k samples for rubric generation and 30k samples for reasoning generation. 

\textbf{Evaluation.} We evaluate on a diverse set of science reasoning benchmarks: 1) Verifiabl: \textsc{GPQA-Diamon} \citep{rein2024gpqa}, \textsc{SciBench} \citep{wang2023scibench}, \textsc{PIQA} \citep{bisk2020piqa}, ResearchQA\citep{yifei2025researchqa}, 2) non-verifiable: \textsc{RaR-Science}~\citep{gunjal2025rubrics} and \textsc{RubricHub}~\citep{li2026rubrichub}. For open-ended scientific tasks, we use gpt-4.1-mini as LLM-as-a-Judge to evaluate following their paper setup and take a 500 subset from the test set. To assess out-of-domain generalization, we further report results on medical question answering benchmarks, including \textsc{MedMCQA} \citep{pal2022medmcqa} and \textsc{PubMedQA} \citep{jin2019pubmedqa}. We follow the \textsc{Language Model Open Science Evaluation} framework for science reasoning.\footnote{\url{https://github.com/GAIR-NLP/lm-open-science-evaluation}}. 
%Details of dataset construction, evaluation settings, and prompting templates are provided in Appendix~\ref{app:eval_details}.

\textbf{Baselines.} We compare the following methods: \emph{(1) supervised fine-tuning}, which trains the student on the distilled CoT trajectories, \emph{(2) GRPO}~\citep{shao2024deepseekmath}, where the reward is a scalar reward produced by an LLM judge, (3) \emph{Rubric-GRPO}, where the reward is aggregated by prompting LLM-as-a-Judge to assign specific rubric rewards, (4) \emph{On-Policy Self-Distillation~\citep{zhao2026opsd}}, where teacher conditions on reference answers and provides dense token-level supervision on student trajectories.
%, \emph{(4) Critique-GRPO~\citep{zhang2025critiquegrpo}:}\footnote{We report performance using the authors' released checkpoint.} a critique-augmented GRPO baseline where the judge produces free-form natural-language critique and append such critique to produce off-policy rollout.

\textbf{Implementation details.} We train with LoRA ($r=64$, $\alpha=128$) and AdamW with learning rate $5\times10^{-6}$,  batch size 32, and FlashAttention 3. Following previous implementations on self-distillation\citep{zhao2026opsd}, we train \method, OPSD for 100 steps and GRPO for 500 steps. More specifically, The rubric generator is trained for 1000 steps with a maximum sequence length of 2,048. The reasoner is trained for 100 steps with a maximum completion length of 4096. The teacher is fixed during training, and supervision is applied through token-level distillation on student-generated trajectories. At evaluation time, we use temperature $1.0$, top-$p=0.95$, top-$k=-1$, min-$p=0.0$, presence penalty $0.0$, and results are averaged over 4 independent generations. For LLM-as-a-Judge evaluation, we report results from a single generation.  Details are in Appendix~\ref{app:impl}.

\subsection{Main Results} 
\begin{table*}[t]

\centering

\small

\setlength{\tabcolsep}{4pt}

\begin{tabular}{lccccccc}
\toprule
\textbf{Method} 
& \textbf{GPQA-D} 
& \textbf{SciBench}
& \textbf{PIQA}
& \textbf{RaR}
& \textbf{ResearchQA}
& \textbf{RubricHub}
& \textbf{Avg} \\
\midrule
Qwen3-8B          & 60.6 & 69.1 & 90.2 & 59.7 & 64.9 & 50.8 & 65.9 \\
+SFT              & 62.0 & 65.4 & 89.2 & 61.5 & 63.2 & 51.2 & 65.4 \\
+GRPO             & 63.5 & 69.7 & 90.3 & 68.2 & 71.5 & 51.9 & 69.2 \\
+GRPO-Rubrics     & 62.1 & 70.2 & 90.3 & \textbf{69.9} & 72.1 & 52.9 & 69.6 \\
+OPSD             & 63.6 & 68.7 & 90.1 & 68.5 & 72.8 & 54.5 & 69.7 \\
\rowcolor{ourblue}
+\method (Ours)      & \textbf{64.5} & \textbf{70.8} & \textbf{90.8} & 68.6 & \textbf{73.1} & \textbf{55.7} & \textbf{70.6} \\
\bottomrule
\end{tabular}
\caption{Main results on diverse reasoning benchmarks. Qwen3-8B is used as the backbone model. GRPO is trained for 500 steps while self-distillation based methods are trained for 100 steps, following prior work's implementation. Higher is better for all metrics.}

\label{tab:main_results}
\end{table*}
Main results are reported in Table~\ref{tab:main_results}. \method achieves the best overall average, improving over the base Qwen3-8B model by 4.7 points and outperforming the strongest baseline, OPSD, by 0.9 point. The gains are especially pronounced on rubric-based reasoning benchmarks, where \method improves over the base model by 8.2 points on \textsc{ResearchQA} and 4.9 points on \textsc{RubricHub}. \method also obtains the best performance on \textsc{SciBench}, reaching 70.8, suggesting that rubric-guided self-distillation remains effective for scientific reasoning tasks that require preserving multiple criterion-level reasoning signals. Compared with GRPO and GRPO-Rubrics, \method benefits from dense token-level supervision rather than relying on scalar rewards. Compared with OPSD, \method avoids distilling from a single reference-style answer and instead uses rubric-conditioned guidance, which appears to provide more flexible and task-relevant supervision. We also observe improvements on \textsc{GPQA-D}, \textsc{PIQA}, \textsc{ResearchQA}, and \textsc{RubricHub}, showing that the method improves not only scientific reasoning but also broader rubric-guided reasoning performance.

\textbf{Generalization to other Domains.} We also evaluate generalization ability on medicine benchmarks in Table~\ref{tab:medicine_results}. Although our method is trained primarily on scientific reasoning tasks rather than medical-domain data, it maintains competitive performance on both \textsc{MedMCQA} and \textsc{PubMedQA}. In particular, \method improves over the base Qwen3-8B model on both benchmarks, from 64.5 to 65.8 on \textsc{MedMCQA} and from 74.2 to 75.1 on \textsc{PubMedQA}. These results suggest that rubric-guided teacher supervision does not lead to catastrophic forgetting on adjacent knowledge-intensive domains, while preserving strong general reasoning ability beyond the training distribution.

\begin{wraptable}{r}{0.5\columnwidth}
\vspace{-0.5em}
\centering
\small
\setlength{\tabcolsep}{4pt}
\begin{tabular}{lcc}
\toprule
\textbf{Method} 
& \textbf{MedMCQA} 
& \textbf{PubMedQA} \\
\midrule
Qwen3-8B        & 64.5 & 74.2 \\
+GRPO   & 65.1 & \textbf{76.2} \\
+GRPO-Rubrics   & 65.6 & 74.2 \\
+OPSD           & \textbf{66.0} & 74.4 \\
\rowcolor{ourblue}
+\method (ours)    & 65.8 & 75.1 \\
\bottomrule
\end{tabular}
\caption{Generalization to medicine benchmarks. Best performance is bolded.}
\label{tab:medicine_results}
\vspace{-1em}
\end{wraptable}

\begin{table*}[t]
\centering
\small
\setlength{\tabcolsep}{4pt}
\begin{tabular}{lccccccc}
\toprule
\textbf{Loss Type}
& \textbf{GPQA-D}
& \textbf{SciBench}
& \textbf{PIQA}
& \textbf{RaR}
& \textbf{ResearchQA}
& \textbf{RubricHub}
& \textbf{Avg} \\
\midrule
Forward KL   & \textbf{64.5} & \textbf{70.6} & \textbf{90.8} & 68.6 & 73.1 & 55.7 & \textbf{70.6} \\
JSD          & 62.1 & 69.2 & 89.9 & 67.1 & 73.1 & 56.2 & 69.6 \\
Reverse KL   & 63.4 & 69.1 & 90.5 & 69.1 & 72.6 & 54.8 & 69.9 \\
\bottomrule
\end{tabular}
\caption{Ablation on the loss type used for \method. All variants use the same Qwen3-8B backbone. Higher is better for all metrics.}
\label{tab:kl_ablation}
\end{table*}

\paragraph{Ablation on Loss Type.}
Table~\ref{tab:kl_ablation} studies the effect of different distillation losses. Forward KL performs best overall, achieving the highest average of 70.6. Reverse KL is the second-best variant, with an average of 69.9, and performs best on \textsc{RaR}. JSD achieves the best result on \textsc{RubricHub} and matches Forward KL on \textsc{ResearchQA}, but shows a noticeable drop on \textsc{GPQA-D}, leading to a lower overall average of 69.6. These results suggest that Forward KL is the most effective objective for \method in this setting. One possible explanation is that Forward KL encourages the student to cover the teacher's rubric-conditioned distribution more faithfully, preserving diverse criterion-aware reasoning signals. In contrast, Reverse KL may be more mode-seeking, while JSD provides a more conservative update that is stable but less effective in transferring the full teacher signal.

\begin{figure}[t]
    \centering
    \includegraphics[width=0.9\textwidth]{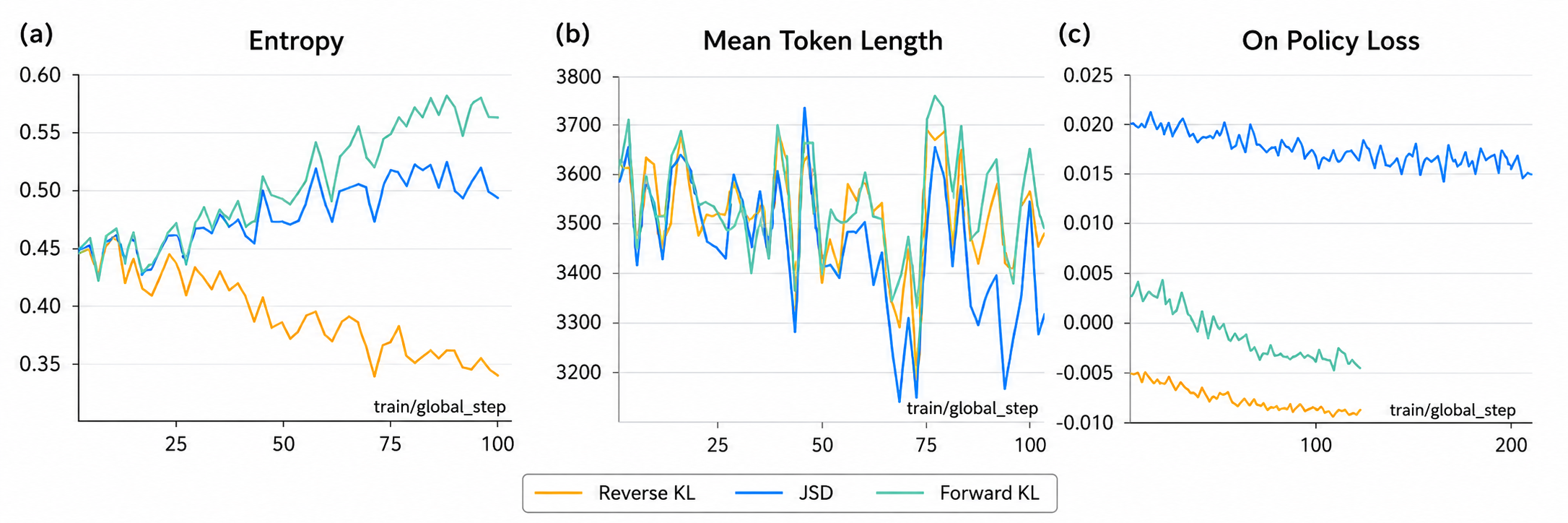}
    \caption{Training dynamics under different distillation losses: reverse KL (yellow), JSD (blue), and forward KL (green). We show student entropy, mean token length, and on-policy loss over training.}
    \label{fig:kl_dynamics}
\end{figure}

Figure~\ref{fig:kl_dynamics} further explains the loss-type ablation in Table~\ref{tab:kl_ablation}. Forward KL achieves the best benchmark performance and shows the most favorable training behavior: student entropy increases steadily, suggesting that the model preserves a broader rubric-conditioned output distribution rather than collapsing to narrow modes. In contrast, reverse KL consistently reduces entropy, reflecting its more mode-seeking tendency, while JSD produces a milder entropy increase. Mean token length is noisy across all objectives, with no clear systematic advantage, although JSD tends to generate slightly shorter responses later in training. The on-policy loss curves are also stable overall: forward KL decreases smoothly, reverse KL becomes increasingly negative, and JSD remains positive with a gradual decline. Together, these trends suggest that forward KL most effectively transfers the teacher's criterion-aware signal, whereas reverse KL is more restrictive and JSD is stable but less effective.

\begin{figure*}[t]
\centering
\begin{subfigure}[t]{0.32\textwidth}
\centering
\includegraphics[width=\linewidth]{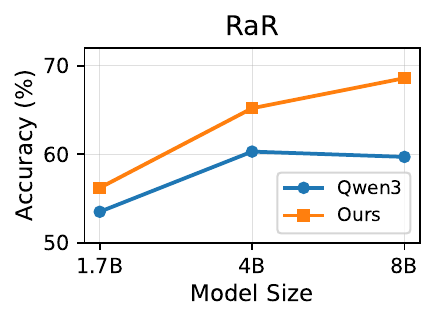}
\end{subfigure}
\hfill
\begin{subfigure}[t]{0.32\textwidth}
\centering
\includegraphics[width=\linewidth]{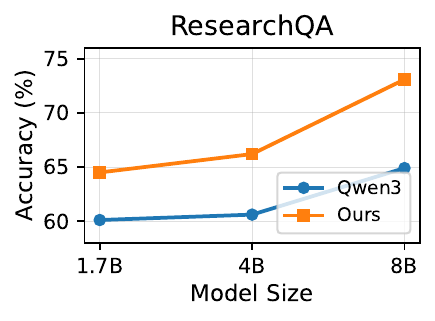}
\end{subfigure}
\hfill
\begin{subfigure}[t]{0.32\textwidth}
\centering
\includegraphics[width=\linewidth]{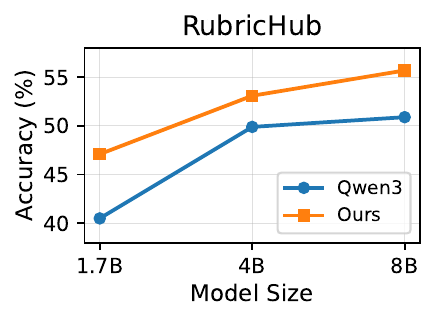}
\end{subfigure}
\caption{Model size ablation on rubric-based reasoning benchmarks at 1.7B, 4B, and 8B scales. \method consistently improves performance over the corresponding base model across \textsc{RaR}, \textsc{ResearchQA}, and \textsc{RubricHub}.}
\label{fig:rubric_model_size}
\end{figure*}

\textbf{Scaling Model Size.} To study whether \method remains effective across scales, we evaluate Qwen3-1.7B, Qwen3-4B, and Qwen3-8B on \textsc{RaR}, \textsc{ResearchQA}, and \textsc{RubricHub}. Figure~\ref{fig:rubric_model_size} shows that \method consistently improves over the corresponding base model across all three benchmarks and model sizes. At 1.7B, \method improves performance by 2.7 points on \textsc{RaR}, 4.4 points on \textsc{ResearchQA}, and 6.6 points on \textsc{RubricHub}. At 4B, the gains are 4.9, 5.6, and 3.2 points, respectively, and at 8B, the gains are 8.9, 8.2, and 4.8 points. These results suggest that \method provides robust gains across model scales, while larger models achieve stronger absolute performance and continue to benefit from criterion-aware teacher signals on rubric-guided reasoning tasks.

\textbf{Ablation on Rubric Source.} Another central question is whether learned rubrics can approach the effectiveness of reference rubrics. Table~\ref{tab:rubric_source} compares \textbf{GT Rubrics}, where the teacher conditions on reference rubrics, with \textbf{Generated Rubrics}, where the teacher conditions on rubrics produced by the learned rubric generator. Overall, generated rubrics remain competitive across the benchmark suite, with only small gaps to GT rubrics on \textsc{GPQA-D} (64.5 vs.\ 65.2), \textsc{SciBench} (70.6 vs.\ 71.0), and \textsc{PIQA} (90.8 vs.\ 91.0). This indicates that the learned rubric generator preserves most of the benefit of reference rubrics. Generated rubrics contain slightly more criteria on average than reference rubrics (8.4 vs.\ 7.5), with a broader range of criterion counts (6--20 vs.\ 7--12), while having a similar average token length (236.6 vs.\ 248.7). This suggests that learned rubrics remain comparably informative for training despite not relying on manually written reference rubrics. This is important in practice: it means we do not rely on handcrafted or manually curated rubric annotations and that the two-stage pipeline is a viable instantiation of \method. Qualitative failure analysis is provided in Appendix~\ref{app:rubric_failure}. Our results further show that the model trained with GT rubrics outperforms all the baselines in Table~\ref{tab:main_results}, demonstrating the effectiveness of token-level distillation on rubric feedback.

\begin{table*}[t]
\centering
\small
\setlength{\tabcolsep}{4pt}
\begin{tabular}{lccccccc}
\toprule
\textbf{Method} 
& \textbf{GPQA-D} 
& \textbf{SciBench}
& \textbf{PIQA}
& \multicolumn{3}{c}{\textbf{Criteria}}
& \textbf{Token Len.} \\
\cmidrule(lr){5-7}
& 
& 
& 
& \textbf{Avg}
& \textbf{Min}
& \textbf{Max}
& \textbf{Avg} \\
\midrule
GT Rubrics        
& \textbf{65.2} & \textbf{71.0} & \textbf{91.0}
& 7.5 & 7 & 12 & 248.7 \\
Generated Rubrics 
& 64.5 & 70.6 & 90.8
& 8.4 & 6 & 20 & 236.6 \\
\bottomrule
\end{tabular}
\caption{Ablation on rubric source. GT (Ground Truth) Rubrics use reference rubrics during reasoner training, while Generated Rubrics correspond to the learned rubric generator.}
\label{tab:rubric_source}
\end{table*}

% \begin{table*}[t]
% \centering
% \small
% \setlength{\tabcolsep}{4pt}
% \begin{tabular}{lccccccc}
% \toprule
% \textbf{Method}
% & \textbf{GPQA-D}
% & \textbf{SciBench}
% & \textbf{PIQA}
% & \textbf{Dataset 1}
% & \textbf{Dataset 2}
% & \textbf{Dataset 3}
% & \textbf{Avg} \\
% \midrule
% Qwen3-8B          & 60.6 & 69.1 & 90.2 & 59.7 & 64.9 & 50.8 & 65.9 \\
% & 61.9 & \textbf{70.9} & 90.2 & -- & -- & -- & 84.0 \\
% \rowcolor{ourblue}
% +100 steps
% & \textbf{64.5} & 70.6 & \textbf{90.8} & -- & -- & -- & \textbf{84.5} \\
% +150 steps
% & 62.4 & 70.7 & 90.5 & -- & -- & -- & 84.2 \\
% +300 steps
% & 63.4 & 70.4 & 90.2 & -- & -- & -- & 84.0 \\
% \bottomrule
% \end{tabular}
% \caption{Main results on diverse reasoning benchmarks with avg@4 and pass@4 ($n{=}4$). Qwen3-8B is used as the backbone. \method checkpoints are from \texttt{qwen3\_8b\_reward\_reasonfirst\_lr5e6\_gen4096\_b0\_rar\_science\_gt\_v2}. Higher is better for all metrics.}
% \label{tab:main_results}
% \end{table*}

\begin{table*}[t]
\centering
\small
\setlength{\tabcolsep}{4pt}
\begin{tabular}{lccccccc}
\toprule
\textbf{Method}
& \textbf{GPQA-D}
& \textbf{SciBench}
& \textbf{PIQA}
& \textbf{RaR}
& \textbf{ResearchQA}
& \textbf{RubricHub}
& \textbf{Avg} \\
\midrule
Qwen3-8B
& 60.6 & 69.1 & 90.2 & 59.7 & 64.9 & 50.8 & 65.9 \\
+8b Direct
& 62.0 & 70.2 & 90.4 & 68.5 & 72.9 & 55.4 & 69.9 \\
+14b Direct
& \textbf{64.9} & 70.6 & 90.1 & \textbf{69.2} & 72.8 & \textbf{56.5} & \textbf{70.7} \\
\rowcolor{ourblue}
+\method (Ours)
& 64.5 & \textbf{70.8} & \textbf{90.8} & 68.6 & \textbf{73.1} & 55.7 & 70.6 \\
\bottomrule
\end{tabular}
\caption{Ablation on the necessity of the Stage-I rubric generator. +14b Direct and +8b Direct prompt Qwen3-14B and Qwen3-8B teachers to provide rubrics directly, while +\method uses rubrics produced by the learned Stage-I rubric generator. }
\label{tab:stage1_rubric_generator}
\end{table*}

\textbf{Necessity of Stage-I Rubric Generator.} We further study whether the training benefits from the learned rubric generator over directly prompting the base model to generate rubrics for teacher model. As shown in Table~\ref{tab:stage1_rubric_generator}, \method achieves a competitive average of 70.6, nearly matching +14b Direct at 70.7 and outperforming +8b Direct at 69.9. Notably, \method obtains the best results on \textsc{SciBench}, \textsc{PIQA}, and \textsc{ResearchQA}, suggesting that the learned generator can amortize rubric construction while preserving strong downstream supervision. Although +14b Direct slightly leads on the overall average, it requires prompting a larger teacher to produce rubrics directly; in contrast, the Stage-I generator provides a scalable way to generate task-specific rubrics for \method.

\begin{table*}
\centering
\small
\setlength{\tabcolsep}{4pt}
\begin{tabular}{lccccccc}
\toprule
\textbf{Method}
& \textbf{GPQA-D}
& \textbf{SciBench}
& \textbf{PIQA}
& \textbf{RaR}
& \textbf{ResearchQA}
& \textbf{RubricHub}
& \textbf{Avg} \\
\midrule
Qwen3-8B
& 60.6 & 69.1 & 90.2 & 59.7 & 64.9 & 50.8 & 65.9 \\
+Generic
& 62.1 & 70.2 & 90.2 & 68.7 & 72.6 & 55.6 & 69.9 \\
+Noisy
& 61.6 & \textbf{70.9} & 90.4 & \textbf{68.8} & 72.4 & 55.3 & 69.9 \\
+Random
& 63.3 & 70.3 & 90.5 & 68.2 & 72.6 & 54.8 & 70.0 \\
+Reduced
& 63.0 & 70.6 & 90.6 & \textbf{68.8} & 72.3 & \textbf{55.7} & 70.2 \\
\rowcolor{ourblue}
+Learned
& \textbf{64.5} & 70.8 & \textbf{90.8} & 68.6 & \textbf{73.1} & \textbf{55.7} & \textbf{70.6} \\
\bottomrule
\end{tabular}
\caption{Ablation on rubric quality. +Generic, +Noisy, +Random, and +Reduced use degraded rubric variants to test the sensitivity of distillation to rubric quality, while +Learned uses the full learned rubric supervision. Bold indicates the best result in each column.}
\label{tab:rubric_degrade}
\end{table*}

\textbf{Analysis of Rubric Quality Degradation.}
We evaluate the sensitivity of \method to rubric quality by providing the teacher with generic, random, noisy, or reduced rubrics. Generic uses a shared rubric template for all examples; Random samples a rubric from another question; Noisy combines half of the original rubric with half of a random rubric; and Reduced removes half of the rubric items. As shown in Table~\ref{tab:rubric_degrade}, all variants improve over the base model, suggesting that criterion-style supervision remains useful even with imperfect rubrics and that the teacher can partially correct unreasonable rubric items. Nevertheless, +Learned achieves the best overall average and the strongest results on \textsc{GPQA-D}, \textsc{PIQA}, and \textsc{ResearchQA}, indicating that instance-specific learned rubrics provide the most reliable supervision. The competitiveness of degraded variants suggests that the teacher can compensate for imperfect rubric information, likely by using the reference answer to reconcile rubric errors. However, the advantage of +Learned shows that rubric relevance and coherence still matter. The strong performance of +Reduced further suggests that \method does not depend on verbose rubrics; concise but relevant criteria preserve most of the benefit.

\section{Related Work}
\paragraph{LLM Post-training}
Reinforcement learning (RL) has become a critical post-training tool for improving multi-step reasoning in LLMs \citep{zhang2025survey}, particularly in the form of \emph{reinforcement learning with verifiable rewards} (RLVR) \citep{shao2024deepseekmath,guo2025deepseek,chollet2025arc,jain2024livecodebench}. Despite these advances, most existing RL methods for reasoning remain fundamentally \emph{outcome-based}. This creates a key limitation for reasoning tasks: the single-score reward is assigned to the entire sequence, and information about \emph{where} and \emph{how} the model failed is lost. Two responses may receive the same reward despite making very different mistake. Recent work moves beyond scalar-only rewards by introducing finer-grained supervision, either through critique-augmented learning that provides natural-language feedback on sampled reasoning traces or through process reward models that assign credit to intermediate steps in a reasoning chain \citep{zhang2025critiquegrpo, bi2025reward, lightman2023lets, setlur2024rewarding, yao2026prl}. Yet these methods are mainly developed for domains such as math, where intermediate steps are easier to evaluate. A complementary line on on-policy distillation replaces scalar outcome rewards with dense teacher guidance. Classical knowledge distillation and sequence-level distillation train a student to imitate a teacher's outputs, but typically operate off-policy on teacher-generated trajectories \citep{hinton2015distilling,kim2016sequence}. Recent work on \emph{on-policy distillation} and \emph{on-policy self-distillation} addresses this mismatch by training the student on trajectories sampled from its own policy while using a teacher to provide token-level supervision \citep{agarwal2024policy,xu2024speculative,zhao2026opsd,hubotter2026sdpo,ye2026opcd}. However, existing on-policy self-distillation methods typically construct the teacher from privileged reference solutions or textual feedback. Our work builds on this line but reshapes the supervision interface: instead of conditioning the teacher on a single reference trajectory, we condition it on structured rubric feedback.

%In GRPO-style training and related approaches, the final learning signal is ultimately reduced to a scalar reward assigned to the sampled response.
%Such approaches mitigate the distribution mismatch of off-policy distillation and the exposure bias of supervised fine-tuning, while potentially improving generalization beyond pure imitation \citep{agarwal2024policy,chu2025sft}.

\paragraph{Reinforcement Learning with Rubrics}
The \emph{LLM-as-a-Judge} paradigm enables scalable evaluation when human labeling is expensive or ambiguous, but coarse holistic scores are often noisy and sensitive to prompting or formatting \citep{zheng2023judging,wang2023pandalm}. Rubric-based evaluation addresses this limitation by decomposing quality into explicit and interpretable criteria, improving consistency and enabling more fine-grained diagnosis \citep{gunjal2025rubrics,arora2025healthbench,starace2025paperbench}. Building on this idea, recent work has incorporated rubrics into reinforcement learning as structured reward decompositions, extending RL-style post-training beyond strictly verifiable domains while providing more interpretable supervision than scalar rewards alone \citep{huang2025reinforcement,gunjal2025rubrics,zhang2025chasing,bi2025reward,DR-Tulu,fang2026rubric}. In these approaches, rubric judgments are aggregated into scalar rewards and applied to completed responses. As a result, rubric structure helps determine \emph{what score} a response receives, but not \emph{how} token-level learning is carried out on the model's own trajectory. Many approaches also rely on predefined rubrics, often produced by frontier LLMs, which are costly to obtain \citep{DR-Tulu}. Our work differs from prior rubric-based RL in the role assigned to rubrics: we use them as privileged teacher-side supervision for on-policy self-distillation, allowing criterion-level structure to directly shape token-level updates during optimization.

%most existing rubric-based training methods still use rubrics primarily at the \emph{evaluation} stage:
%
%More importantly, rubric-driven supervision for intermediate reasoning remains relatively underexplored, even though many failures arise from flawed reasoning processes rather than only the final answer itself. This observation is consistent with recent findings that structured process-level feedback, such as critique-augmented learning, can improve exploration and robustness compared with purely outcome-based rewards \citep{zhang2025critiquegrpo, bi2025reward}. 
% \input{chapters/05-conclusion}

\section{Conclusion}
We introduced \emph{Rubric-Conditioned Self-Distillation}, a post-training framework that uses rubrics as privileged teacher supervision for on-policy self-distillation. Instead of collapsing rubric feedback into scalar rewards, \method preserves criterion-level structure during optimization by converting rubrics into dense token-level guidance on student-generated rollouts. Empirically, \method shows that preserving criterion-level feedback during on-policy distillation leads to stronger and more fine-grained reasoning. More broadly, \method offers a verifier-free approach to post-training open-ended reasoning models, where high-quality responses cannot always be judged by exact-match answers, executable tests, or other automatic outcome verifiers.

\bibliography{colm2026_conference}
\bibliographystyle{colm2026_conference}
\newpage

\appendix
\section{Implementation Details}

%\subsection{Implementation Details}
In \textbf{standard GRPO} with an LLM judge, each rollout is scored by a fixed judge model (Qwen3-14B), which is conditioned on the user prompt, the model completion, and a reference answer. The judge outputs a discrete quality score on a 1–10 scale, which we normalize to ($[0,1]$) to serve as the reward signal. In \textbf{GRPO-Rubrics} method, we first perform an offline rubric-generation step, where Qwen3-14B, prompted as a rubric writer, maps each training example’s question and reference answer into a structured rubric. During training, the reward judge is conditioned on the prompt, the model completion, and the pre-generated rubric, and produces a holistic score. As a result, rewards reflect rubric satisfaction rather than direct comparison to the gold text. Training configurations are reported in Table~\ref{tab:training_config}.
\label{app:impl}
\begin{table}[H]
\centering
\caption{Training Configurations}
\label{tab:training_config}
\small
\setlength{\tabcolsep}{8pt}
\renewcommand{\arraystretch}{1.15}
\begin{tabular}{lccc}
\toprule
\textbf{Parameter} & \textbf{GRPO} & \textbf{\method/OPSD} & \textbf{SFT} \\
\midrule
Learning Rate & $5 \times 10^{-6}$ & $5 \times 10^{-6}$ & $5 \times 10^{-6}$  \\
Max Completion Length & 4096 & 4096 & 4096\\
 Batch Size & 32 & 32 & 32\\
Sampling Temperature & 1.2 & 1.2 & 1.2\\
Training Steps & 500 & 100 & 500 \\
Number of Generations per Prompt & 4 & 1  & 1\\
\bottomrule
\end{tabular}
\end{table}

% \subsection{Accuracy-Efficiency Tradeoff.} \method exhibits best accuracy-efficiency tradeoff overall: It is both more accurate and more token-efficient across all datasets. In contrast, Qwen3-8B, GRPO, and GRPO-Rubric generally operate in a higher-token regime, and these longer outputs do not consistently translate into better accuracy.

\section{Prompt Templates}
\label{app}

We provide the prompt templates used in our rubric generation and LLM-as-a-Judge evaluation, where we adopt the prompt format from RaR~\citep{gunjal2025rubrics}. For rubric generation, we instruct the model to generate self-contained, criterion-level rubrics in a structured JSON format. For LLM-as-a-Judge evaluation, we ask the evaluator to holistically score a generated response according to the provided rubrics and return a JSON-formatted rating.

\subsection{Rubric Generator Prompt}
\label{app:rubric}

\begin{tcblisting}{
title=Rubric Generator Prompt,
colback=blue!3,
colframe=blue!35!black,
breakable,
enhanced,
listing only,
listing options={
basicstyle=\ttfamily\footnotesize,
breaklines=true,
columns=fullflexible
}
}
You are an expert rubric writer. Your job is to generate a self-contained set of evaluation criteria ("rubrics") for judging how good a response is to a given question. Rubrics can cover aspects of a response such as, but not limited to, factual correctness, ideal-response characteristics, style, completeness, helpfulness, harmlessness, patient-centeredness, depth of reasoning, contextual relevance, and empathy.

Each item must be self-contained: non-expert readers should not need to infer anything or consult external information. Begin each description with its category: "Essential Criteria: ...", "Important Criteria: ...", "Optional Criteria: ...", or "Pitfall Criteria: Does not mention ...".

Inputs:

question: The full question text.
reference_answer: The ideal answer, including any specific facts, explanations, or advice.

Total items:

Choose 5--15 rubric items based on the complexity of the question.

Each rubric item:

title: 2--4 words.
description: One sentence starting with its category prefix that explicitly states exactly what to look for.
weight: For Essential/Important/Optional, use 1--5; for Pitfall, use -1 or -2.

Category guidance:

Essential: Critical facts or safety checks; if missing, the response is invalid. Use weight 5.
Important: Key reasoning, completeness, or clarity; strongly affects quality. Use weight 3--4.
Optional: Helpful style or extra depth; nice to have but not deal-breaking. Use weight 1--2.
Pitfall: Common mistakes or omissions specific to this prompt. Use weight -1 or -2.

To ensure self-contained guidance:

When referring to answer choices, explicitly say "Identifies (A)", "Identifies (B)", etc., rather than vague phrasing.
If the format requires a conclusion like "The final answer is (B)", include a rubric item such as: "Essential Criteria: Includes a clear statement 'The final answer is (B)'."
If reasoning should precede the answer, include a rubric item such as: "Important Criteria: Presents the explanation before stating the final answer."
If brevity is valued, include a rubric item such as: "Optional Criteria: Remains concise and avoids unnecessary detail."
If the question context demands mention of specific findings, include that explicitly.

Output:
Provide a JSON array of rubric objects. Each object must contain exactly three keys: title, description, and weight. Do not copy large blocks of the question or reference_answer into the text. Each description must begin with its category prefix, and no extra keys are allowed.

Now, given the question and reference_answer, generate the rubric as described. The reference answer is an ideal response but not necessarily exhaustive; use it only as guidance.
\end{tcblisting}

\subsection{LLM-as-a-Judge Prompt}
\label{app:judge_prompt}

\begin{tcblisting}{
title=LLM-as-a-Judge Prompt,
colback=blue!3,
colframe=blue!35!black,
breakable,
enhanced,
listing only,
listing options={
basicstyle=\ttfamily\footnotesize,
breaklines=true,
columns=fullflexible
}
}
System Prompt:
You are an expert evaluator. Given a user prompt, a generated response, and a list of quality rubrics, please rate the overall quality of the response on a scale of 1 to 10 based on how well it satisfies the rubrics.

Consider all rubrics holistically when determining your score. A response that violates multiple rubrics should receive a lower score, while a response that satisfies all rubrics should receive a higher score.

Start your response with a valid JSON object that starts with json and ends with . The JSON object should contain a single key "rating" and the value should be an integer between 1 and 10.

Example response:

{
  "rating": 7
}

User Prompt Template:
Given the following prompt, response, and rubrics, please rate the overall quality of the response on a scale of 1 to 10 based on how well it satisfies the rubrics.

Your JSON Evaluation:
\end{tcblisting}

\section{Experiments}

\begin{figure}[t]
\centering
\includegraphics[width=0.45\columnwidth]{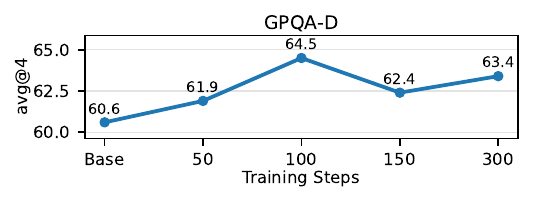}
\vspace{-0.5em}
\includegraphics[width=0.45\columnwidth]{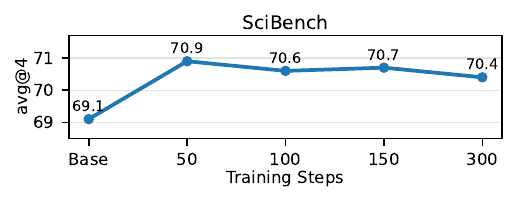}
\vspace{-0.8em}
\caption{Performance across training checkpoints on GPQA-D and SciBench.}
\label{fig:checkpoint_curves}
\end{figure}

\textbf{Training Steps Analysis.} Figure~\ref{fig:checkpoint_curves} shows the performance trajectory across different training checkpoints. On \textsc{GPQA-D}, performance improves steadily from the base model and peaks at 120 steps, suggesting that \method is most effective with moderate training. On \textsc{SciBench}, performance improves early and remains relatively stable across later checkpoints, indicating that the method quickly adapts to scientific reasoning tasks without substantial degradation from continued training.

\subsection{Case Study on Reasoning Quality}
\label{sec:qualitative_analysis}

We compare model responses on a simple ideal-gas-law problem. Although all responses identify the correct governing equation \(P=\frac{nRT}{V}\), they differ substantially in verbosity, consistency, and numerical discipline.

\paragraph{Prompt.}
\emph{Suppose that \(10.0\,\mathrm{mol}\ \mathrm{C_2H_6(g)}\) is confined to \(4.860\,\mathrm{dm}^3\) at \(27^\circ\mathrm{C}\). Predict the pressure exerted by the ethane from the perfect gas. The unit of the answer is \(\mathrm{atm}\). Let's think step by step.}

\begin{tcolorbox}[title=\method, colback=blue!3, colframe=blue!35!black, breakable]
\small
\textbf{Token count:} 2896

\textbf{Selected quotes:}
\begin{quote}
``First, I need to convert the temperature from Celsius to Kelvin. The given temperature is \(27^\circ\mathrm{C}\). To convert, I add 273.15. So \(27 + 273.15 = 300.15\,\mathrm{K}\).''

``The volume is given in \(\mathrm{dm}^3\), which is the same as liters ... \(V = 4.860\,\mathrm{L}\).''

``Since the answer should be in atm, I need to use the value of \(R\) that has units \(\mathrm{L\cdot atm/(mol\cdot K)}\) ... \(R = 0.08206\).''

``Now plug these into \(PV=nRT\). Solving for \(P\): \(P = \frac{nRT}{V}\).''

``So total \(P \approx 50.679647\). Rounding to three decimal places gives \(50.680\,\mathrm{atm}\).''

``The answer is therefore \(\boxed{50.680}\).''
\end{quote}

\textbf{Response excerpt.}
The response first converts temperature to Kelvin, maps \(4.860\,\mathrm{dm}^3\) to \(4.860\,\mathrm{L}\), selects \(R=0.08206\,\mathrm{L\cdot atm/(mol\cdot K)}\), and applies
\[
P=\frac{nRT}{V}.
\]
It then computes
\[
P=\frac{10.0\times 0.08206\times 300.15}{4.860}
=\frac{246.30309}{4.860}\approx 50.680.
\]
The response ends with a short, clean answer:
\[
\boxed{50.680}.
\]
\end{tcolorbox}

\begin{tcolorbox}[title=OPSD, colback=green!3, colframe=green!35!black, breakable]
\small
\textbf{Token count:} 4936

\textbf{Selected quotes:}
\begin{quote}
``The ideal gas law is \(PV=nRT\), right?''

``Since the answer needs to be in atmospheres, I should use the \(R\) value that has units of \(\mathrm{L\cdot atm/(mol\cdot K)}\) ... \(R = 0.082057\).''

``The formula is \(P = nRT / V\).''

``So the numerator is approximately \(246.2940855\,\mathrm{L\cdot atm}\).''

``So total is approximately \(50.677 + 0.000795 \approx 50.6778\). So approximately \(50.6778\,\mathrm{atm}\). Rounding to three decimal places, that would be \(50.678\,\mathrm{atm}\).''

``Therefore, the answer is \(\boxed{50.678}\).''
\end{quote}
\textbf{Response excerpt.}
The response follows the same overall structure, but uses a more precise constant \(R=0.082057\,\mathrm{L\cdot atm/(mol\cdot K)}\), leading to
\[
P=\frac{10.0\times 0.082057\times 300.15}{4.860}
=\frac{246.2940855}{4.860}\approx 50.677795.
\]
After repeated recalculation and self-verification, it rounds to
\[
\boxed{50.678}.
\]
The response is correct, but noticeably more repetitive than \method.
\end{tcolorbox}

\begin{tcolorbox}[title=Qwen3-8B, colback=red!3, colframe=red!35!black, breakable]
\small
\textbf{Token count:} 9294

\textbf{Selected quotes:}
\begin{quote}
``The gas constant \(R\) is \(0.0821\,\mathrm{L\cdot atm/(mol\cdot K)}\), right?''

``Wait, but sometimes \(R\) is taken as \(0.08206\) ...''

``The exact value of \(R\) is \(0.082057\) ...''

``So with \(R = 0.0821\), the answer is approximately \(50.704\,\mathrm{atm}\).''

``But if I use \(R = 0.08206\), it's approximately \(50.680\,\mathrm{atm}\).''

``But since the problem might expect using \(R = 0.0821\), I think the answer is expected to be around \(50.704\,\mathrm{atm}\).''
\end{quote}

\textbf{Response excerpt.}
The response begins correctly by identifying \(PV=nRT\), converting \(27^\circ\mathrm{C}\) to \(300.15\,\mathrm{K}\), and noting that \(4.860\,\mathrm{dm}^3=4.860\,\mathrm{L}\). However, it repeatedly switches between different values of the gas constant:
\[
R=0.0821,\quad 0.08206,\quad 0.082057.
\]
This causes the intermediate calculations to drift between
\[
50.704,\quad 50.680,\quad 50.678,
\]
and the response becomes very long, circular, and self-contradictory. In particular, the model explicitly states that it will ``proceed with \(R=0.0821\),'' which yields approximately
\[
\boxed{50.704}.
\]
Even though the underlying formula is correct, the response demonstrates weaker numerical stability and substantially poorer reasoning efficiency.
\end{tcolorbox}

This example reveals a clear difference in \emph{reasoning quality}, not just final-answer correctness. All three models know the ideal gas law and the required unit conversions, but they differ in how efficiently and consistently they execute the solution. \method produces the strongest trajectory. Its response is the shortest, commits early to a coherent numerical setup, and reaches a correct answer without revisiting earlier choices. \textsc{OPSD} also arrives at a correct answer, but its reasoning is more verbose and repetitive: it repeatedly re-checks arithmetic that has already been established, which increases token usage without improving solution quality. In contrast, the base \textsc{Qwen3-8B} response is substantially longer and exhibits a more serious failure mode: it repeatedly changes core numerical assumptions, especially the value of \(R\), and consequently oscillates between different final answers. Overall, this case study suggests that \method improves \emph{reasoning efficiency} and \emph{trajectory stability}: its responses are shorter, less repetitive, and more internally consistent than those produced by OPSD and the base model.

\subsection{Failure Analysis on Learned Rubric Quality}
\label{app:rubric_failure}

\begin{tcolorbox}[title=Learned Rubrics, colback=red!3, colframe=red!35!black, breakable]
\small
\textbf{Phonon / Bose--Einstein Example}
\begin{enumerate}
    \item \textbf{Bose-Einstein Derivation} (5): Essential Criteria: The response must clearly derive the Bose-Einstein distribution formula for phonon occupation number $n(\omega)$ using the given dispersion relation $\omega(k)=ck$.
    \item \textbf{Temperature Dependence} (4): Important Criteria: The response should correctly express the temperature dependence of $n(\omega)$ as $n(\omega)=1/(e^{\hbar \omega / kT}-1)$, demonstrating the exponential dependence on temperature and frequency.
    \item \textbf{Frequency-Dependent Behavior} (5): Essential Criteria: The response must explicitly explain how the occupation number $n(\omega)$ changes with temperature for different $\omega$ values, emphasizing the inverse relationship between $n(\omega)$ and temperature.
    \item \textbf{Physical Interpretation} (4): Important Criteria: The response should include a clear explanation of the physical meaning of the Bose-Einstein distribution in the context of phonon statistics and thermal equilibrium.
    \item \textbf{High/Low Temperature Behavior} (3): Optional Criteria: The response may include a brief discussion on the behavior of $n(\omega)$ at high and low temperatures, such as the classical limit or zero-point energy effects.
    \item \textbf{Frequency Dependence Omission} (-1): Negative Criteria: The response must not omit the dependence of the occupation number on the frequency $\omega$, as this is a critical aspect of the temperature dependence.
    \item \textbf{Temperature Independence Error} (-1): Negative Criteria: The response must not incorrectly state that the occupation number is independent of temperature, which would contradict the Bose-Einstein distribution.
    \item \textbf{Dispersion Relation Usage} (4): Important Criteria: The response should correctly use the dispersion relation $\omega(k)=ck$ to connect the phonon frequency with wavevector $k$ in the derivation.
    \item \textbf{Bose-Einstein Application} (4): Important Criteria: The response must demonstrate the correct application of Bose-Einstein statistics to phonons, including the proper handling of the occupation number formula.
    \item \textbf{Frequency-Temperature Interaction} (3): Optional Criteria: The response may include a clear explanation of how the occupation number changes with frequency for different temperature values, enhancing the understanding of the system.
\end{enumerate}

\textbf{Soda-Lime Titration Example}
\begin{enumerate}
    \item \textbf{Mass Conversion} (5): Essential Criteria: The response must accurately convert the mass of soda lime to the masses of NaOH and CaO components using the given percentages.
    \item \textbf{Mole Calculation} (5): Essential Criteria: The response must correctly compute the moles of NaOH and CaO from their respective masses using their molar masses.
    \item \textbf{Reaction Equations} (4): Important Criteria: The response should accurately identify and write the balanced chemical equations for the neutralization reactions of NaOH and CaO with HCl.
    \item \textbf{Stoichiometric Ratio} (4): Important Criteria: The response must correctly apply stoichiometric ratios from the balanced equations to relate moles of NaOH and CaO to moles of HCl required for neutralization.
    \item \textbf{Total Moles Calculation} (5): Essential Criteria: The response should correctly calculate the total moles of HCl required by summing the moles from both NaOH and CaO neutralization steps.
    \item \textbf{Volume Calculation} (5): Essential Criteria: The response must accurately determine the volume of 0.500M HCl needed using the total moles and the given molarity, converting to the correct units.
    \item \textbf{Step-by-Step Explanation} (4): Important Criteria: The response should present a clear, step-by-step explanation of the calculation process for both NaOH and CaO neutralization, ensuring logical flow.
    \item \textbf{Application Context} (3): Optional Criteria: The response may include a brief discussion of the significance of the neutralization reactions in real-world applications, though it is not required.
    \item \textbf{Component Omission} (-1): Negative Criteria: The response must not omit the neutralization of either NaOH or CaO components, as both are required for accurate calculation.
    \item \textbf{Molar Mass Accuracy} (-1): Negative Criteria: The response must not use incorrect molar masses for NaOH or CaO, as this would lead to wrong results.
\end{enumerate}
\end{tcolorbox}

\begin{tcolorbox}[title=Reference Rubrics (RaR-Science), colback=blue!3, colframe=blue!35!black, breakable]
\small
\textbf{Phonon / Bose--Einstein Example}
\begin{enumerate}
    \item \textbf{Bose-Einstein Distribution} (5): Essential Criteria: The response must explicitly state and correctly use the Bose-Einstein distribution formula for $n(\omega)$, such as $n(\omega)=1/(\exp(\hbar\omega/(k_B T))-1)$, linking $\omega$ and $T$ in the derivation.
    \item \textbf{Dispersion Relation Use} (4): Important Criteria: The answer should correctly incorporate the given dispersion relation $\omega(k)=ck$ to connect the frequency $\omega$ to the wave vector $k$ in the context of the phonon system.
    \item \textbf{Temperature Analysis} (5): Essential Criteria: The response must analyze how the phonon occupation number $n(\omega)$ varies with temperature for a fixed frequency and explain differences in behavior at various $\omega$ values.
    \item \textbf{Mathematical Derivation} (4): Important Criteria: The answer should include clear and logically structured derivations that break down the mathematical steps required to arrive at the temperature dependence of $n(\omega)$.
    \item \textbf{Frequency Trends} (3): Optional Criteria: The response may provide a concrete example or detailed explanation illustrating that at lower frequencies the occupation number is more sensitive to changes in temperature than at higher frequencies.
    \item \textbf{Clarity and Conciseness} (3): Optional Criteria: The answer should be clear and concise, avoiding unnecessary elaboration while still covering all key elements of the derivation and conclusion.
    \item \textbf{Exclusion of Loss Effects} (-1): Pitfall Criteria: The response should not include irrelevant factors such as frictional or damping losses, which are not part of the ideal derivation using Bose-Einstein statistics.
\end{enumerate}

\textbf{Soda-Lime Titration Example}
\begin{enumerate}
    \item \textbf{Separate Reactions} (5): Essential Criteria: The response must separately address the neutralization reactions for both NaOH and CaO components, calculating the moles of acid required for each reaction.
    \item \textbf{Stoichiometry Accuracy} (5): Essential Criteria: The answer should correctly apply stoichiometric relationships to determine the moles of HCl needed for the complete neutralization of both compounds.
    \item \textbf{Molarity Application} (4): Important Criteria: The response must demonstrate how the molarity of 0.500M HCl is used to convert the required moles of acid into the corresponding volume in cm$^3$ with proper unit conversions.
    \item \textbf{Step-by-Step Work} (4): Important Criteria: The answer should provide a clear, logical sequence of calculations that lead to the final volume, ensuring transparency in each intermediary step.
    \item \textbf{Final Volume Accuracy} (5): Essential Criteria: The response must explicitly state the correct final volume of 0.500M HCl required (133.04 cm$^3$) for complete neutralization.
    \item \textbf{Unit Consistency} (2): Optional Criteria: The explanation should include correct unit conversions, especially showing how volumes are converted (e.g., L to cm$^3$), to enhance clarity and precision.
    \item \textbf{Reaction Assumptions} (-2): Pitfall Criteria: The response should mention that the reactions are assumed to go to completion without interference, and neglecting to state such assumptions is a common oversight.
\end{enumerate}
\end{tcolorbox}

\paragraph{Failure analysis.}
These examples illustrate two recurring weaknesses of the learned rubrics relative to the reference rubrics. First, the learned rubrics are often \emph{bloated and redundant}. In the phonon/Bose--Einstein example, the learned rubric expands to 10 criteria versus 7 in the reference rubric, and several titles repeat the same underlying evaluation axis: \emph{Temperature Dependence}, \emph{Frequency-Dependent Behavior}, \emph{High/Low Temperature Behavior}, and \emph{Frequency-Temperature Interaction} all partially restate the same requirement. By contrast, the reference rubric compresses this content into a smaller and sharper set of criteria, such as \emph{Temperature Analysis}, \emph{Mathematical Derivation}, and \emph{Frequency Trends}. This suggests that the learned rubric generator tends to over-segment closely related concepts instead of consolidating them into a compact set of discriminative checks.

Second, the learned rubrics sometimes include \emph{generic but weakly task-critical criteria}. In the soda-lime titration example, the learned rubric includes items such as \emph{Step-by-Step Explanation} and \emph{Application Context}, which are broadly reasonable but not central to the actual grading target. The reference rubric instead concentrates on the chemically essential checks: \emph{Separate Reactions}, \emph{Stoichiometry Accuracy}, \emph{Final Volume Accuracy}, and \emph{Unit Consistency}. Overall, these cases suggest that the learned rubrics usually identify the correct topic, but they are often more verbose, overlapping, and less precisely aligned with the true grading objective than the reference rubrics.

\section{Additional Theoretical Details}
\label{app:theory}

\paragraph{Proposition 1 (Robustness to rubric approximation).}
Let \(r^\star\) be a reference rubric and \(\hat r\) a generated rubric. For each visited prefix \((x,\hat y_{<t})\), define
\[
p_t^\star := p_T^Y(\cdot \mid x,r^\star,\hat y_{<t}),
\qquad
\hat p_t := p_T^Y(\cdot \mid x,\hat r,\hat y_{<t}),
\qquad
q_t := p_S^Y(\cdot \mid x,\hat y_{<t}).
\]
Assume that \(-\log q_t(y)\le B_t\) for all tokens \(y\) in the support of \(p_t^\star\) and \(\hat p_t\). Then
\[
\Big|
\mathbb{E}_{y\sim p_t^\star}[-\log q_t(y)]
-
\mathbb{E}_{y\sim \hat p_t}[-\log q_t(y)]
\Big|
\le
2 B_t\, \mathrm{TV}(p_t^\star,\hat p_t),
\]
where \(\mathrm{TV}\) denotes total variation distance. Consequently,
\[
\big|
\mathcal{L}_{\mathrm{CE}}(\hat r)-\mathcal{L}_{\mathrm{CE}}(r^\star)
\big|
\le
\mathbb{E}_{\hat y \sim p_S^Y(\cdot \mid x)}
\left[
\sum_{t=1}^{|\hat y|}
2 B_t\, \mathrm{TV}(p_t^\star,\hat p_t)
\right].
\]

\paragraph{Proof.}
Fix a visited prefix \((x,\hat y_{<t})\), and define
\[
f_t(y):=-\log q_t(y).
\]
By assumption, \(f_t(y)\le B_t\) for all \(y\) in the support of \(p_t^\star\) and \(\hat p_t\). Then
\begin{align}
&\Big|
\mathbb{E}_{y\sim p_t^\star}[f_t(y)]
-
\mathbb{E}_{y\sim \hat p_t}[f_t(y)]
\Big| \nonumber \\
&=
\left|
\sum_y \bigl(p_t^\star(y)-\hat p_t(y)\bigr) f_t(y)
\right| \nonumber \\
&\le
\sum_y \bigl|p_t^\star(y)-\hat p_t(y)\bigr|\, |f_t(y)| \nonumber \\
&\le
B_t \sum_y \bigl|p_t^\star(y)-\hat p_t(y)\bigr| \nonumber \\
&=
2 B_t\, \mathrm{TV}(p_t^\star,\hat p_t).
\end{align}
Applying this bound at each decoding step, summing over \(t\), and taking expectation over student rollouts \(\hat y \sim p_S^Y(\cdot \mid x)\) yields the stated result. \qed

Proposition~1 shows that generated rubrics do not need to match reference rubrics exactly in wording or surface form. It is sufficient that they induce similar rubric-conditioned teacher distributions on the prefixes actually visited by the student. This matches the practical role of Stage~I: the rubric generator needs to produce rubrics that are useful enough to preserve the teacher-side guidance used in Stage~II.

% \section{LLM Usage Disclosure}
% We used LLMs for proofreading, grammar correction, and sentence refinements. These tools were not used to generate core scientific ideas, results, or conclusions.

\end{document}